\documentclass[sigconf]{acmart}

\AtBeginDocument{%
  }

\copyrightyear{2023}
\acmYear{2023}
\setcopyright{licensedusgovmixed}\acmConference[GeoAI '23]{6th ACM SIGSPATIAL International Workshop on AI for Geographic Knowledge Discovery}{November 13, 2023}{Hamburg, Germany}
\acmBooktitle{6th ACM SIGSPATIAL International Workshop on AI for Geographic Knowledge Discovery (GeoAI '23), November 13, 2023, Hamburg, Germany}
\acmPrice{15.00}
\acmDOI{10.1145/3615886.3627747}
\acmISBN{979-8-4007-0348-5/23/11}

\usepackage{subcaption}
\usepackage{makecell}
\usepackage{multirow}

\begin{document}

\title{Assessment of a new GeoAI foundation model for flood inundation mapping}

\author{Wenwen Li}
\authornote{Corresponding Author}
\orcid{1234-5678-9012} 
\affiliation{%
  \institution{School of Geographical Sciences and Urban Planning, Arizona State University}
  \city{Tempe}
  \state{AZ}
  \country{USA}
  \postcode{85287-5302}
}
\email{wenwen@asu.edu}

\author{Hyunho Lee}
\affiliation{%
  \institution{School of Geographical Sciences and Urban Planning, Arizona State University}
  \city{Tempe}
  \state{AZ}
  \country{USA}
  \postcode{85287-5302}
  }
\email{hlee401@asu.edu}

\author{Sizhe Wang}
\affiliation{%
  \institution{School of Computing and Augmented Intelligence, Arizona State University}
  \city{Tempe}
  \state{AZ}
  \country{USA}}
  \postcode{85287-5302}
\email{wsizhe@asu.edu}

\author{Chia-Yu Hsu}
\affiliation{%
  \institution{School of Geographical Sciences and Urban Planning, Arizona State University}
  \city{Tempe}
  \state{AZ}
  \country{USA}}
  \postcode{85287-5302}
\email{chsu53@asu.edu}

\author{Samantha T. Arundel}
\affiliation{%
  \institution{United States Geological Survey, Center of Excellence for Geospatial Information Science (CEGIS)}
  \city{Rolla}
  \state{MO}
  \country{USA}}
  \postcode{65401}
\email{sarundel@usgs.gov}

\renewcommand{\shortauthors}{Li, et al.}

\begin{abstract}

Vision foundation models are a new frontier in Geospatial Artificial Intelligence (GeoAI), an interdisciplinary research area that applies and extends AI for geospatial problem solving and geographic knowledge discovery,  because of their potential to enable powerful image analysis by learning and extracting important image features from vast amounts of geospatial data. This paper evaluates the performance of the first-of-its-kind geospatial foundation model, IBM-NASA’s Prithvi, to support a crucial geospatial analysis task: flood inundation mapping. This model is compared with convolutional neural network and vision transformer-based architectures in terms of mapping accuracy for flooded areas. A benchmark dataset, Sen1Floods11, is used in the experiments, and the models' predictability, generalizability, and transferability are evaluated based on both a test dataset and a dataset that is completely unseen by the model. Results show the good transferability of the Prithvi model, highlighting its performance advantages in segmenting flooded areas in previously unseen regions. The findings also indicate areas for improvement for the Prithvi model in terms of adopting multi-scale representation learning, developing more end-to-end pipelines for high-level image analysis tasks, and offering more flexibility in terms of input data bands.


\end{abstract}



\maketitle

\section{Introduction}

Floods are one of the most devastating natural disasters, affecting over 20\% of the world's population \citep{rentschler2022flood}. Recent climate change and extreme weather events have led to an increase in both the frequency and intensity of flooding, resulting in significant economic losses and loss of life \citep{li2015performance}. To effectively conduct risk assessments and disaster management, flood inundation mapping—an effort to accurately delineate the extent of flooding—has become a vital tool. Flood maps are not only used by insurance companies to assess flood risks but are also integral to early warning systems, aiding residents in preparing for and evacuating from potential flooding events \citep{li2023geographvis}. Additionally, flood maps assist engineers and city planners in designing resilient infrastructure to mitigate flood-related impacts \citep{li2020real}.

\begin{figure*}[t]
  \centering
  \includegraphics[width=1.0 \linewidth]{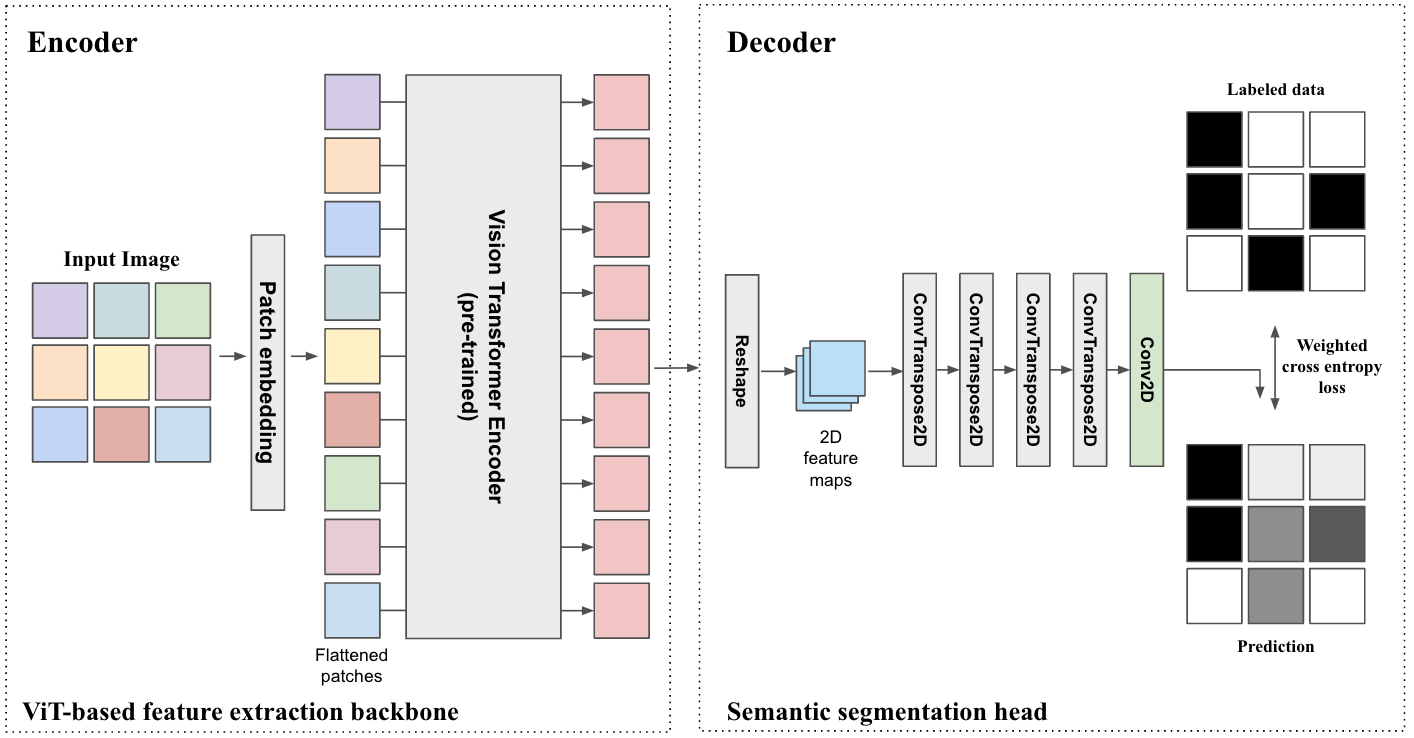}
  \caption{The architecture of the geospatial foundation model Prithvi, tailored for semantic segmentation.}
  \Description{}
  \label{fig:arch-of-gfm}
\end{figure*}

Traditional methods for flood inundation mapping rely on a combination of field surveys, aerial photography, topographic maps, hydrological models, and historical data. While providing very valuable information for risk assessment and mitigation planning, these methods are often costly and the data are time-consuming to collect. In recent years, advanced remote sensing platforms, such as Sentinel and Planet, are generating finer-resolution satellite imagery with high revisit frequency and multiple spectral bands, and have become important data sources for flood mapping. In addition to the advances in data, the rapid development of GeoAI technology\citep{li2020geoai}, especially deep learning, also offers a more informed and automated approach to analyze big satellite data \citep{hsu2023explainable, hsu2021knowledge, hu2019geoai, li2021tobler, li2022real, wang2021geoai}. Through data-driven analysis, deep learning models have demonstrated the capability to recognize distinctive flood patterns and subtle flood boundaries, even in regions under complex and rapidly changing conditions. 

In GeoAI, flood inundation mapping can be formulated as a semantic segmentation problem, which aims to provide per-pixel classification of flood and non-flood areas. Popular deep learning methods that can tackle this problem include U-Net, a convolutional neural network (CNN) architecture that uses a CNN-based encoder for feature extraction and a decoder for the reconstruction of image features at the original resolution by considering spatial contextual information \citep{ronneberger2015u}. This encoder-decoder based architecture in U-Net has been proven very effective in image segmentation and it has become the foundation for many semantic segmentation models, such as ResUNet \citep{diakogiannis2020resunet} and SegNet \citep{badrinarayanan2017segnet}. Another family of convolutional semantic segmentation models include the DeepLab family developed by Google \citep{chen2017deeplab}. The DeepLab family uses dilated convolutions to enable multi-scale feature aggregation, and the model has found many applications in geospatial problems. However, for segmenting floods, which are less frequently occurring events than some of the others, the benchmark training datasets are relatively small in size. As a result, DeepLab did not show a performance as high as expected on such applications due to its model complexity and the needs of more training data \citep{yadav2022deep}.  

Recently, the transformer-based architecture has shown even stronger performance than CNN-based models due to its capability in capturing data dependencies over long ranges and the ability to reduce inductive bias \citep{li2022geoai}. Transformer was originally designed for natural language processing, and then adopted in many image analysis and vision tasks through the development of vision transformers (ViT). For such dense prediction (e.g., per-pixel) as semantic segmentation, ViT needs to be extended to add a segmentation head. Commonly used models of this category include Mask2Former \citep{cheng2022masked}, Swin transformer \citep{liu2021swin}, and Segformer \citep{xie2021segformer}. Like U-Net, these models often adopt an encoder-decoder based architecture, but with different designs in the attention mechanism and mask segmentation. Segformer, for example, consists of a hierarchical transformer encoder and a lightweight decoder based on a multilayer perceptron. This model uses a smaller patch size than regular ViT-based models to achieve precise segmentation. Its efficient architecture also yields fast processing speed while achieving competitive segmentation results on multiple benchmark datasets \citep{thisanke2023semantic}.

A new wave of AI research is the development of large foundation models \citep{bommasani2021opportunities}. They are models trained on huge amounts of data leveraging often self-supervised learning to achieve unprecedented generalizability for downstream tasks. In the computer vision domain, notable progress has been made. For instance, Meta released the Segment Anything Model (SAM) \citep{kirillov2023segment} in April 2023 to achieve zero-shot instance segmentation with prompts. This model was trained on 11 million images and 1 billion masks, which were created by human annotation, and then semi-automated and fully-automated learning. Although SAM provides fine-grained, instance-level segmentation, it cannot be easily adopted for semantic segmentation due to the differences in the goals of these tasks and the training data used. In the GeoAI domain, a billion-scale foundation model was developed \citep{cha2023billion}. This model is built with 2.4 billion parameters, its backbone is also transformer-based architecture, and it is pre-trained on a remote sensing benchmark dataset, the MillionAID \citep{long2021creating}. However, the model is not open-sourced; therefore, its adaptability in different domain tasks is difficult to assess. In August 2023, IBM and NASA jointly released a new geospatial foundation model (GFM), Prithvi. This model was trained using self-supervised learning on Harmonized Landsat and Sentinel 2 (HLS) data within the continental United States. Because both datasets are satellite imagery capturing important spectral signatures of the Earth surface, the foundation model learned from them tends to gain more geospatial domain knowledge than models trained from other natural images. Hence, it has high potential to better support geospatial tasks. 

\begin{figure*}[t]
  \centering
  \includegraphics[width=1.0 \linewidth]{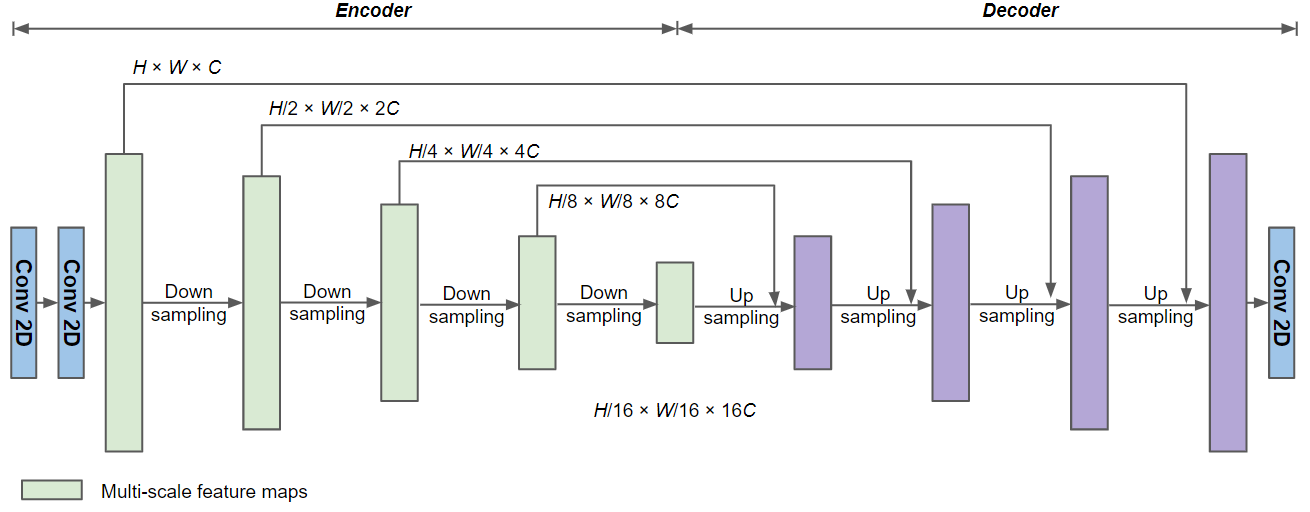}
  \caption{Architecture of U-Net. H: height. W: width. C: channel.}
  \Description{}
  \label{fig:arch-of-unet}
\end{figure*}

This paper aims to assess IBM and NASA's new foundation model in support of flood inundation mapping. We compare the model with commonly used CNN and transformer-based semantic segmentation architectures, especially U-Net and Segformer, in their predictability, generalizability, and transferability. The goal is to better understand the strengths and weaknesses of this new solution and how it can be best leveraged to support disaster assessment related to flood. 

The rest of the paper is organized as follows. Section 2 will introduce Prithvi’s architecture, along with the other two comparative models. Section 3 describes the experimental settings, including dataset, training parameters, and evaluation metrics. Section 4 presents the results and analysis. Section 5 concludes the work and proposes future research directions. 


\section{Models}

\subsection{IBM-NASA’s geospatial foundation model - Prithvi} 

IBM-NASA's GFM Prithvi \citep{Prithvi-100M} adopts a temporal ViT trained on 30-m HLS data covering the continental United States. The model pre-training is based on a Masked AutoEncoder (MAE) \citep{he2022masked} strategy, which asks the model to learn and predict a masked patch from the original image through self-supervised learning. Different from the general ViT model, which is trained on single time slice images, the Prithvi model (Figure \ref{fig:arch-of-gfm}, encoder part) can take in time-series satellite images with an additional time dimension, and the patch embedding process is customized such that both spatial and positioning information as well as the temporal information are captured in the resulting embeddings. After that, the embeddings are sent to the transformer encoder for capturing inherent spatial and temporal dependencies within the datasets, creating a set of features similar to the feature map obtained from a CNN-based backbone. 

It is important to note that the original decoder of the model used for MAE-based pre-training might not be suitable for various downstream tasks. To better support image analysis tasks, a task-specific head, such as a segmentation head, needs to be added to build the analytical pipeline. Figure \ref{fig:arch-of-gfm} presents the integrated semantic segmentation pipeline for flood inundation mapping. The encoder is a pre-trained module of the model and it works as described above. The decoder is responsible for pixel-wise image segmentation and it works as follows. First, the encoded vectors will be deserialized to create two-dimensional (2D)-shaped feature maps. To improve the segmentation precision, the feature map with reduced size as compared to the original image will go through an upsampling process, which goes through four 2D transposed convolutional layers. Finally, a 2D convolutional layer is added to the end of the decoder to make the final predictions. In the flood mapping case, there are three classes to predict: flood, non-flood, and no data. 

An important feature of the Prithvi model is that its input requires six bands, rather than the typical three RGB (red, green, blue) bands, used in other models. These six bands include: red, green, blue, narrow NIR (Near InfraRed), SWIR1 (Short-Wave Infrared) and SWIR2. SWIR1 and SWIR2 differ in their wavelengths, and they are used for discriminating moisture content (SWIR1) and improved moisture content (SWIR2) of soil and vegetation \citep{loveland2016landsat}.  

\begin{figure*}[t]
  \centering
  \includegraphics[width=0.8 \linewidth]{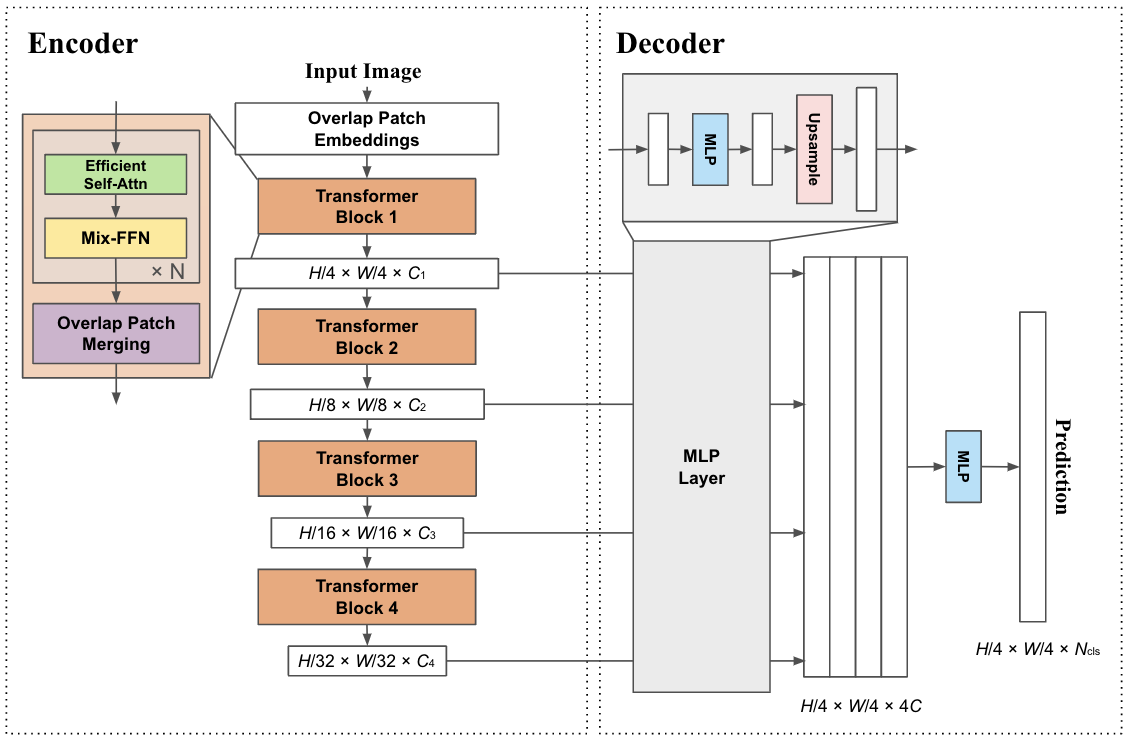}
  \caption{Model architecture of Segformer (Adapted from \citep{xie2021segformer}). H: height. W: width. C: channel.}
  \Description{}
  \label{fig:arch-of-segformer}
\end{figure*}


\subsection{Comparative models for flood inundation mapping}
\subsubsection{U-Net} 

The U-Net model is selected for performance comparison with Prithvi because it is a classic model for semantic segmentation \citep{ronneberger2015u} and one of the most popular models adopted for flood inundation mapping \citep{konapala2021exploring}. U-Net is a convolutional neural network-based architecture, consisting of two parts: an encoder and a decoder (Figure \ref{fig:arch-of-unet}). The encoder provides a contracting path, where important image features are extracted hierarchically through convolution. The size of the feature map also gradually decreases by using downsampling to reduce spatial resolution and thereby the overall computational cost. The U-Net encoder used in our experiments contains four downsampling steps, accomplished by two convolutional layers for feature extraction and one max-pooling layer for downsampling. The output of the encoder is fed into the decoder part, which contains an expanding path to recover the resolution of the feature maps to the original image resolution. This is achieved through four upsampling modules (consisting of an interpolation upsampling layer and one convolution layer) to reconstruct spatial details. In addition to its classic encoder-decoder-based architecture, another key design feature of U-Net is the inclusion of skip connections. These connections link the decoder outputs with corresponding features generated during the encoding phase, further enhancing precision in segmentation.


\subsubsection{Segformer} 

Segformer \citep{xie2021segformer} also adopts an encoder-decoder architecture (Figure \ref{fig:arch-of-segformer}) but different from U-Net, it uses a transformer-based feature extraction backbone. In the encoder stage, Segformer introduces hierarchical feature representation to generate CNN-like multi-level features. These feature maps retain both high-level features at low-resolution and low-level features at high-resolution. Segformer enhances computational efficiency through Efficient Self-Attention in its transformer blocks (Figure \ref{fig:arch-of-segformer}). It is a modified self-attention mechanism that scales down the number of input sequences, effectively alleviating the primary computational bottleneck within the encoder part. The model also incorporates a data-driven positional encoding through Mix-FFN (Feed Forward Network), which uses a 3-by-3 convolutional layer within the FFN to encode positional information. This adaptation mitigates the performance drop caused by positional encoding when the test image resolution changes during inference. Segformer also partitioned the images into overlapping patches to preserve the data continuity and local contexts. These patches are merged together before being sent to the next transformer blocks. 

The Segformer decoder, which benefits from the large effective receptive field (ERF) of the transformer encoder, fuses these multi-level features using a lightweight All-MLP (multi-layer perceptron) module. This module exclusively consists of MLP components for generating the segmentation mask, thereby substantially reducing computational demands compared to other models, such as SEgmentation TRansformer (SETR) \citep{zheng2021rethinking}, that utilize multiple convolution layers in its decoder.


\section{Data and experiments}

\subsection{Data}

Sen1Floods11 \citep{bonafilia2020sen1floods11} is utilized to conduct the experiments. Sen1Floods11 is a georeferenced dataset for flood inundation mapping. This dataset contains 446 pairs of Sentinel-1 and Sentinel-2 satellite imagery, which captured 11 flood events from 11 countries (Bolivia, Ghana, India, Nigeria, Pakistan, Paraguay, Somalia, Spain, Sri Lanka, United States, and Vietnam) between 2016 and 2019, with corresponding human expert labeled data. Each satellite image maintains a spatial resolution of 10m with an image size of 512 × 512 pixels. In particular, the Sentinel-2 satellite imagery includes 13 spectral bands at varying resolutions, and all bands are linearly interpolated to 10m resolution to align spatially. Data from 6 Sentinel-2 bands (red, green, blue, narrow NIR, SWIR1, SWIR2) are used as the input to run all three models as they match exactly with the required input of Prithvi. 


\subsection{Experimental setup}

Our experiments evaluate the GFM (Prithvi)’s performance when applied to downstream geospatial tasks, especially flood inundation mapping using semantic segmentation, compared to transformer-based backbone not specifically designed for geospatial data (Segformer using a pretrained model), and a traditional encoder-decoder architecture built from scratch (U-Net). The experiments were conducted based on the MMSegmentation framework \citep{mmseg2020}. For comparison, the experimental setups followed the GFM's configuration for the flood inundation mapping use case provided by Hugging Face (https://huggingface.co/ibm-nasa-geospatial/Prithvi-100M-sen1floods11, accessed on 31 August 2023). 

One exception is the use of the optimizer. The original Prithvi model uses Adam as the optimizer, but we found through our experiments that AdamW \citep{loshchilov2017decoupled}, which uses a different strategy for weight decay, offers better segmentation performance for Prithvi. Hence, AdamW is applied to all three of the comparatives. Additionally, regarding learning rate, we referred to the previous studies of U-Net for flood inundation mapping \citep{konapala2021exploring, bonafilia2020sen1floods11}. As U-Net has a simpler architecture than Segformer and Prithvi, using a slightly larger learning rate helps the model to converge faster.  The detailed experimental settings are described in Table \ref{tab:parameters}. “Poly” in Table \ref{tab:parameters} refers to the learning rate scheduler that linearly reduces the learning rate from the initial value to zero as the training advances. 

\begin{table}[h]
  \caption{The parameters of the deep learning models for the experiment.}
  \label{tab:parameters}
  \begin{tabular}{lccc}
    \toprule
    Parameters & U-Net & Segformer & Prithvi \\
    \midrule
    Optimizer & AdamW & AdamW & AdamW \\
    Learning rate & 5e-4 & 6e-5 & 6e-5 \\
    Weight decay & 0.01 & 0.01 & 0.01 \\
    Batch size & 4 & 4 & 4 \\
    \makecell[l]{Learning rate \\ scheduler} & Poly & Poly & Poly \\
    \makecell[l]{Class weight \\ (flood:non-flood)} & 7:3 & 7:3 & 7:3 \\
    Loss function & \makecell{Cross \\ Entropy} & \makecell{Cross \\ Entropy} & \makecell{Cross \\ Entropy} \\
    \makecell[l]{Cropped image \\ size} & 244 & 244 & 244 \\
    \makecell[l]{Datasets for \\ pretraining} & - & ADE20K \citep{zhou2017scene} & HLS \\
  \bottomrule
\end{tabular}
\end{table}

In terms of architecture, these models exhibit both shared and distinct features. First, both the Prithvi and the Segformer models adopt transformer based architectures, which are known to be powerful but also require a huge amount of data to achieve outstanding performance. Hence, pre-training is often necessary to ensure their performance when being fine-tuned for downstream tasks. In comparison, U-Net is a classic CNN-based model that involves less computation and therefore it can often be trained from scratch. Hence, Table \ref{tab:parameters} shows the pre-training datasets for Prithvi and Segformer. For Segformer, we adopted the model pretrained on a benchmark semantic segmentation dataset ADE20K \citep{zhou2017scene} and the Prithvi is pre-trained on NASA’s HLS dataset. Second, both U-Net and Segformer introduce multi-scale learning in their model design, ensuring a good model performance and high-precision segmentation. Third, U-Net and Segformer are trained based on supervised learning, whereas the pre-training phase of Prithvi uses self-supervised learning and only adopts supervised learning for the fine-tuning part. 

In the experiments, we utilized two types of datasets for performance evaluation. The first consists of test data, which include images from the same geographic region (e.g., country) as the training data but differ from the training images. The second dataset comprises entirely `unseen' data from Bolivia, which was not part of the training, validation, or test datasets. For both datasets, we conducted 10 runs of each model and recorded the performance metrics for each run. These metrics include Intersection over Union (IoU) for each class, mean IoU (mIoU) across all classes, Accuracy (Acc) for each class, and Mean Accuracy (mAcc) across all classes.

Equations 1-4 provide their definitions, which have been adopted from the MMSegmentation framework (TP: True Positive; FN: False Negative; FP: False Positive). It is important to note that the accuracy formula in MMSegmentation is designed for multi-class segmentation, equivalent to the recall formula used in binary segmentation.

\begin{table*}[h]
  \caption{Performance evaluation results on test data. Avg.: Average. Acc: Accuracy. mIoU: mean Intersection over Union. M: million}
  \label{tab:perf-test}
  \begin{tabular}{@{}cccccccc@{}}
    \toprule
    \multirow{2}{*}{{Model}} &
    \multirow{2}{*}{{Avg. mIoU (\%)}} &
    \multicolumn{2}{c}{{Avg. IoU (\%)}} &
    \multirow{2}{*}{\begin{tabular}[c]{@{}c@{}}Avg. mAcc\\ (\%)\end{tabular}} &
    \multicolumn{2}{c}{{Avg. Acc (\%)}} &
    \multirow{2}{*}{{\makecell{Number of trainable \\ parameters}}} \\ \cmidrule(lr){3-4} \cmidrule(lr){6-7}
                                     &                & Flood          & Non-flood      &                & Flood          & Non-flood      &      \\ \midrule
    IBM-NASA’s Prithvi                        & 89.59          & 81.98          & 97.21          & 94.35          & 90.12          & 98.58          & 100M \\
    Segfomer-B0   & 89.36          & 81.57          & 97.14          & 94.20          & 89.84          & 98.55          & 3.7M \\
    Segfomer-B5 & 89.54          & 81.89          & 97.18          & 94.44          & 90.35          & 98.52          & 82M  \\
    U-Net                                & \textbf{90.80} & \textbf{84.03} & \textbf{97.57} & \textbf{94.80} & \textbf{90.74} & \textbf{98.86} & 29M  \\ \bottomrule
\end{tabular}
\end{table*}

\begin{table*}[h]
\caption{Performance evaluation results on the unseen, Bolivia data}
\label{tab:perf-unseen}
\begin{tabular}{@{}cccccccc@{}}
\toprule
\multirow{2}{*}{{Model}} &
  \multirow{2}{*}{{Avg. mIoU (\%)}} &
  \multicolumn{2}{c}{{Avg. IoU (\%)}} &
  \multirow{2}{*}{\begin{tabular}[c]{@{}c@{}}Avg. mAcc\\ (\%)\end{tabular}} &
  \multicolumn{2}{c}{{Avg. Acc (\%)}} &
  \multirow{2}{*}{{\makecell{Number of trainable \\ parameters}}} \\ \cmidrule(lr){3-4} \cmidrule(lr){6-7}
                                     &                & Flood          & Non-flood      &                & Flood          & Non-flood      &      \\ \midrule
IBM-NASA’s Prithvi                        & \textbf{86.02} & \textbf{76.62} & \textbf{95.43} & \textbf{90.38} & \textbf{82.12} & 98.65          & 100M \\
Segfomer-B0  & 83.05          & 71.46          & 94.65          & 86.95          & 74.77          & 99.14          & 3.7M \\
Segfomer-B5 & 85.59          & 75.82          & 95.36          & 89.58          & 80.26          & 98.91          & 82M  \\
U-Net                                & 82.54 & 70.57 & 94.52 & 86.45 & 73.73 & \textbf{99.18} & 29M  \\ \bottomrule
\end{tabular}
\end{table*}

The final result is presented as the average of each metric obtained from the 10 experiments.

\begin{align}
IoU = \frac{TP}{TP + FN + FP} 
\end{align}

\begin{align}
mIoU = \frac{IoU_{flood} + IoU_{nonflood}}{2} 
\end{align}

\begin{align}
Acc = \frac{TP}{TP+FN} 
\end{align}

\begin{align}
mAcc = \frac{Acc_{flood} + Acc_{nonflood}}{2} 
\end{align}

\section{Results and analysis}

Table \ref{tab:perf-test} lists the experimental results from the Prithvi model, the Segformer model, and the U-Net model. Two Segformer models (B0 and B5) with different model sizes were run with the results reported. B0 is the smallest Segformer model and B5 is the largest Segformer model. Their trainable parameters can be found in Table \ref{tab:perf-test}. The experimental results show that the U-Net model obtains the best overall segmentation performance in all measures. This is attributed to U-Net’s unique encoder-decoder architecture and its ability to consider multi-scale features to reconstruct spatial details for high-precision segmentation. 

The performance of Prithvi and the Segformer models is quite similar in terms of IoU, with a 1\% difference in mIoU and nearly a 3\% difference in IoU when segmenting flood pixels, compared to U-Net. The prediction accuracy (mAcc) of Segformer-B5 is slightly higher than that of Prithvi (Table \ref{tab:perf-test}), and the two models have no obvious performance advantage over the Segformer-B0 model. 

In segmenting flood pixels, Segfomer-B5 and Prithvi obtained slightly better accuracy (1.2-1.5\% in Avg. Acc) than the other two models. This indicates these large models can capture diverse spectral characteristics of flood pixels. When we compare the model size with regard to the number of training parameters, Segfomer-B0 only contains 3.7M parameters (Table \ref{tab:perf-test}), which is much smaller than the other models, and it still achieves comparable results. This has largely benefited from its lightweight model design and the use of hierarchical feature fusion in its learning process.

However, the models’ performance for unseen regions are quite different from that in the test regions. Table \ref{tab:perf-unseen}'s results show that Prithvi yields the highest mIoU and prediction accuracy (mAcc) among all comparative models. The second best model is Segformer-B5, with a difference of less than 1\% compared to Prithvi. Segformer-B0 follows, while U-Net ranks lowest in terms of performance, with a nearly 4\% gap in both mIoU and mAcc compared to Prithvi. This result indicates that large models, such as Prithvi and Segfomer-B5, have a greater capability to learn important feature representations, thus possessing stronger domain adaptability and transferability. The geospatial foundation model Prithvi, pre-trained on remote sensing imagery, captures more domain-specific geospatial knowledge, making it particularly suitable for geospatial tasks. Smaller models, such as U-Net, work well in the test data, meaning that the model learns and fits very well with the flood datasets; however, this near “perfect” fitting on a given dataset also limits its ability to adapt to other datasets and tasks. 

\begin{figure*}[t]
  \centering
  \includegraphics[width=1.0\linewidth]{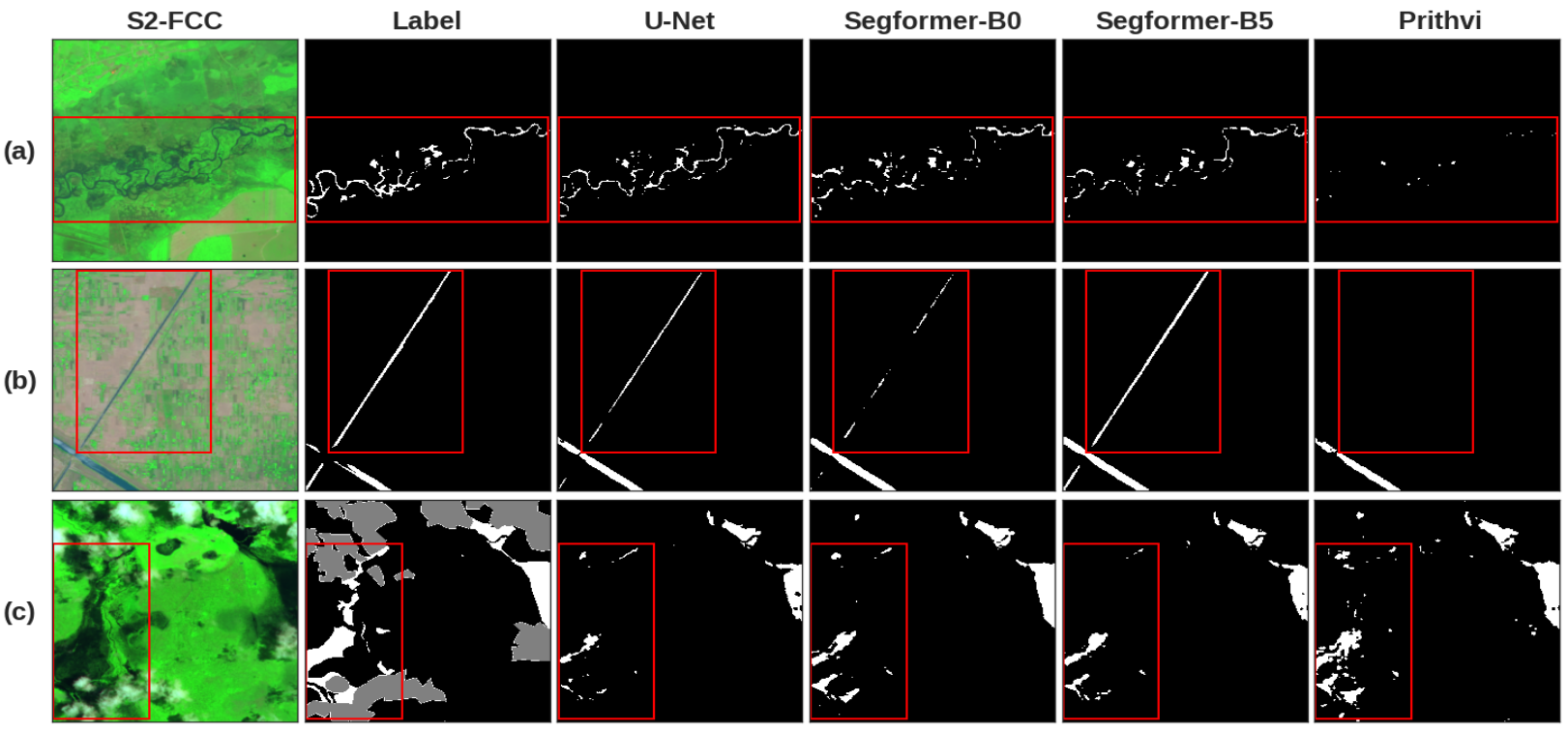}
  \caption{Visual comparison of prediction results. The images in rows (a) and (b) are from the test dataset, and the image in row (c) is from the unseen Bolivia dataset. "Label" indicates ground-truth labels. White: flood; Black: non-flood; Gray: no data. S2-FCC denotes Sentinel-2 False Color Composite. The red boxes highlight the comparative regions in terms of model segmentation results.}
  \Description{}
  \label{fig:pred-image}
\end{figure*}

To further examine the characteristics of each model, we visualized the prediction results in Figure \ref{fig:pred-image}. For the prediction results on the test data (rows (a) and (b)), which contain narrow flood inundation areas (e.g., rivers, channels), the Prithvi model did not perform as well as the other models. This is because the U-Net and Segformer architectures enable multi-level feature extraction, whereas Prithvi only utilizes a single-level ViT-based architecture as the feature extraction backbone. On the other hand, in the Bolivia data, Prithvi demonstrates consistently higher performance than U-Net. The bottom row (c) of Figure \ref{fig:pred-image} shows the segmentation results in the unseen region. It can be seen that Prithvi predicted better results than the other models, especially in the highlighted image regions (red boxes). This is attributed to the model's pre-training on huge amounts of data to achieve good generalizability and, therefore, transferability compared to models learned from and (over)fitted on a small dataset. 

\section{Conclusion}
Research on the GeoAI foundation model has attracted substantial interest in the geospatial community. To analyze Earth and environmental changes, satellite and other remotely sensed images have become crucial data sources \citep{arundel2020geonat,li2023geoimagenet}. To process vast amounts of imagery data, vision foundation models have been developed to parse and understand image scenes, improving image analysis and pattern recognition. Although several such foundation models, such as Meta’s SAM and Microsoft’s Florence, have been developed, they were trained on natural images acquired from optical cameras. These images exhibit different characteristics compared to satellite imagery, which captures changes on the Earth's surface. IBM and NASA have jointly developed a first-of-its-kind geospatial foundation model, Prithvi, trained on large satellite images. This model offers the potential to support a wide variety of geospatial analysis tasks, including image classification, object detection, and image segmentation.

This paper provides an assessment of the Prithvi model regarding its support for flood inundation mapping, which requires the model to provide per-pixel segmentation of its flood and non-flood classes. This model is compared with popular architectures, including (1) U-Net, a classic semantic segmentation model built upon CNN and an encoder-decoder architecture, and (2) Segformer, a ViT-based semantic segmentation model that implements an efficient architecture through a lightweight decoder and multi-level feature extraction. The models are tested on a benchmark dataset, Sen1Floods11, and several findings are identified: (1) the foundation model Prithvi outperforms the other models in segmenting images in unseen regions. This performance is likely attributed to Prithvi's pre-training on a massive set of time-series satellite imagery, reflecting its good model generalizability and transferability, which are desirable properties for GeoAI models \citep{anselin2014metadata,goodchild2021replication,kedron2021reproducibility,wilson2021five}. (2) However, the Prithvi model did not perform as well as the classic U-Net and Segformer models on testing datasets where different images from the same geographical regions have been used for model training. This indicates an area for improvement that would be beneficial for Prithvi, as it currently only uses single-level features for image analysis, whereas the classic U-Net and the new Segformer model both adopt multi-scale features in their model learning process. (3) Unlike other general-purpose foundation models, Prithvi requires three additional bands beyond RGB as model input. This may pose challenges for leveraging image data that do not contain the required bands and therefore negatively affect the model’s predictive performance. (4) The Prithvi model does not include an end-to-end architecture for high-level image analysis tasks, such as object detection or semantic segmentation. Hence, to fully utilize its pre-trained encoder, more components, such as decoder heads, ideally would be added. This may limit its adoption by scientists without essential AI knowledge and skills. Nonetheless, as a new development in GeoAI, the open-source Prithvi model offers great potential for establishing a research community dedicated to advancing GeoAI vision foundation models collaboratively. 

In the future, we plan to conduct a more comprehensive evaluation of the GFM’s performance, architecture, and learning strategies. This could help us fully understand the model’s strengths and weaknesses, contributing to its continuous advancement as a general-purpose foundation model. Ideally, it would support advanced image analysis and vision tasks through simple fine-tuning, few-shot, and zero-shot learning. This way, we anticipate broader adoption across various geospatial applications, further enhancing our scientific understanding of the Earth and its changing climate and environment.

\begin{acks}
This work is supported in part by the National Science Foundation under awards 2120943, 1853864, and 2230034, as well as Google.org grants. 

Any use of trade, firm, or product names is for descriptive purposes only and does not imply endorsement by the U. S. Government.

\end{acks}

\bibliographystyle{ACM-Reference-Format}
\bibliography{2023_acmgis_flood-reference}

\end{document}